%% file: main.tex
\documentclass{article}

\PassOptionsToPackage{numbers,compress}{natbib}
\usepackage[main, final]{neurips_2026}

\usepackage[utf8]{inputenc} 
\usepackage[T1]{fontenc}    
\usepackage{hyperref}       
\usepackage{url}            
\usepackage{booktabs}       
\usepackage{amsfonts}       
\usepackage{nicefrac}       
\usepackage{microtype}      
\usepackage[table]{xcolor}         

\usepackage{graphicx}
\usepackage{subcaption}
\usepackage{amsmath}
\usepackage{amssymb}
\usepackage{dsfont}
\usepackage{algorithm}
\usepackage{algorithmic}
\usepackage{multirow}
\usepackage{array}
\usepackage{caption}
\usepackage{listings}
\usepackage[frozencache]{minted}
\usepackage{proof}

\title{A Principled Optimal Transport Framework for Frame Selection in Long Video Understanding}

\author{%
  Ziwei Zhao \\
  Yizhun Medical AI Co., Ltd \\
  Beijing, China \\
  \texttt{zhaozw998@gmail.com} \\
  \And
  Yuhang Liu \\
  Yizhun Medical AI Co., Ltd \\
  Beijing, China \\
  \texttt{yuhang.liu@yizhun-ai.com}
}

\begin{document}

\maketitle

\begin{abstract}
Long-video understanding with multimodal LLMs is fundamentally constrained by the mismatch between video length and the model's limited visual context budget, making frame selection essential for practical reasoning. Existing query-aware methods typically rely on heuristic pipelines: they assign one-dimensional query-conditioned scores to frames and then apply sampling, ranking, or extraction rules to form the final subset. While effective in practice, such methods are generally not derived from an explicit optimization objective and struggle to capture the heterogeneous evidence required for complex reasoning. In this paper, we revisit long-video frame selection from a principled optimization perspective. We formulate it as a structured evidence-allocation problem and propose \textbf{OTFS}, an optimal transport framework that maps multiple evidence sources onto video frames. To reflect the asymmetry of this problem---important evidence should be preserved, whereas redundant frames need not absorb mass---we instantiate this view with a semi-unbalanced entropic optimal transport objective and an efficient solver, \emph{OTFS-Sinkhorn}. Combined with an exact dynamic program for coverage-aware subset extraction, OTFS yields a practical two-stage training-free pipeline. We further show that, in the single-source case, our formulation induces a Gibbs-type distribution over frames, making Q-Frame's temperature-scaled distribution a special case of our framework while also providing a broader perspective for understanding related prior methods. Experiments on three long-video understanding benchmarks show that OTFS consistently outperforms strong training-free baselines. Codes are available at \href{https://github.com/barry664/OTFS}{https://github.com/barry664/OTFS}.
\end{abstract}

\section{Introduction}

A central challenge in long-video understanding with multimodal large language models (MLLMs) is that long videos contain far more visual content than current models can process within their limited visual context budget~\cite{maaz2023videochatgpt,lin2023videollava,cheng2024videollama,li2024llavaonevision,zhang2024longcontexttransfer,bai2025qwen25vl,zhang2023videollama,lin2023mmvid,ren2024timechat,zohar2024apollo,zhang2025videollama3,tong2024cambrian1,shi2024eagle}. This mismatch between video length and model capacity makes visual compression necessary in practical systems~\cite{song2023moviechat,liu2024kangaroo,weng2024longvlm,qian2024videostreaming,shen2024longvu,xu2024slowfastllava,li2025videochatflash,wang2024retake,zhao2024needlevideo,zhu2025internvl3}. Among existing compression strategies, frame selection---that is, choosing a small subset of frames from the original video---has become a widely used approach. Because only the selected frames remain available to the downstream reasoner, frame selection largely determines which visual evidence can be preserved for long-video question answering and temporal reasoning.

\begin{figure}[t]
  \centering
  \includegraphics[width=\textwidth]{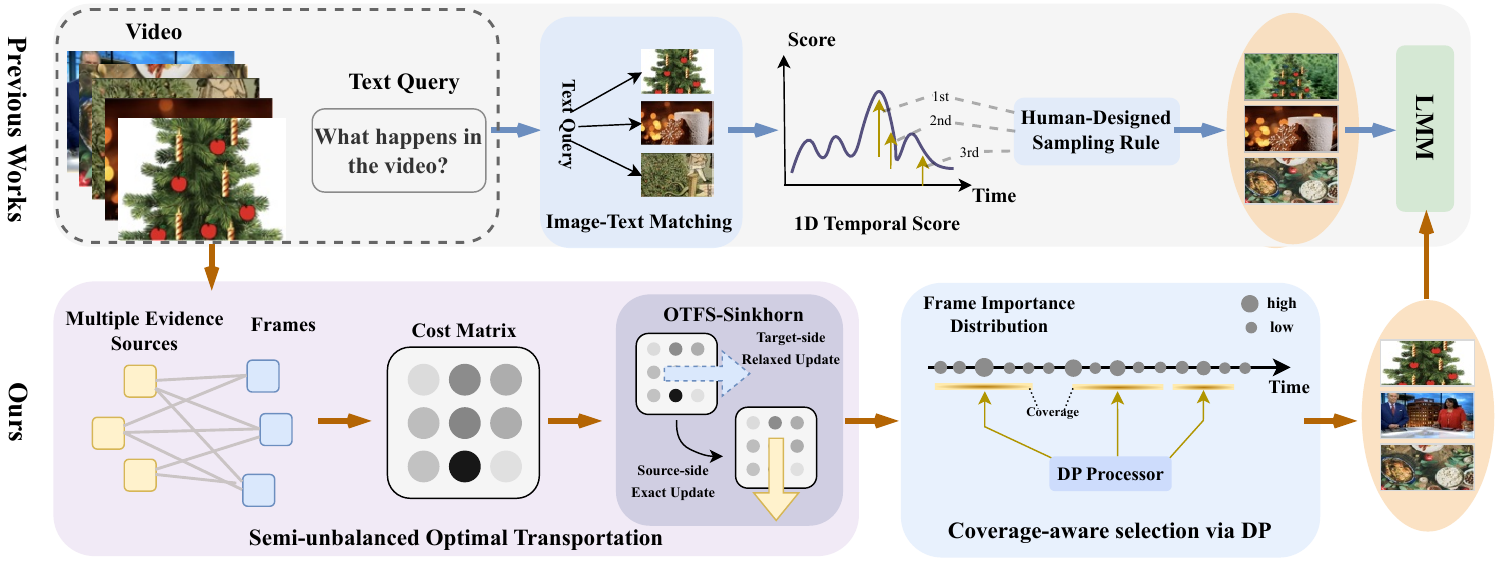}
  \caption{Comparison between prior training-free frame-selection procedures and our principled OT-based framework. Existing methods typically compute one-dimensional frame scores and then apply hand-designed sampling or extraction rules. In contrast, OTFS formulates frame selection as semi-unbalanced transport from evidence atoms to video frames, producing a soft frame-importance distribution that is converted into a coverage-aware subset by exact dynamic programming.}
  \label{fig:intro}
\end{figure}

Existing query-aware frame selection methods have substantially improved over uniform sampling, yet most still rely on heuristic frame-scoring and subset-construction procedures~\cite{tang2025adaptive,zhang2025qframe,zhu2026focus,li2025divide}. As illustrated at the top of Figure~\ref{fig:intro}, they typically derive a one-dimensional query-conditioned score over frames and then apply hand-designed sampling or extraction rules to form the final subset. Although Q-Frame~\cite{zhang2025qframe}, AKS~\cite{tang2025adaptive}, and DIG~\cite{li2025divide} differ in their scoring signals and extraction strategies, \textbf{they are generally not derived as exact solvers of a clearly specified frame-selection objective}. As a result, it remains difficult to reason about optimality under an explicit mathematical target.

This limitation becomes more pronounced in complex long-video reasoning, where answering a question may require multiple types of visual evidence, such as direct semantic relevance, answer-oriented utility, and temporal continuity cues~\cite{tang2025adaptive,li2025divide}. In this setting, what matters is not only how important each frame appears in isolation, but how a limited frame budget should be allocated so that the selected subset collectively preserves the heterogeneous evidence required for downstream reasoning. This suggests viewing frame selection not merely as heuristic frame ranking, but as a structured evidence-allocation problem under budget constraints.

In this paper, we revisit long-video frame selection from a principled optimization perspective and propose \textbf{OTFS}, an \textbf{O}ptimal \textbf{T}ransport framework for \textbf{F}rame \textbf{S}election. Our key idea is to formulate frame selection as transporting evidence requirements onto video frames. Concretely, we represent query- or visually-induced cues as \emph{evidence atoms} on the source side and video frames as candidate locations on the target side. The resulting transport plan captures how different evidence atoms allocate support to different frames, and its target marginal serves as a soft frame-importance distribution. In this way, frame importance is grounded in an explicit transport objective rather than query-aware scores alone.

A key observation is that this allocation problem is intrinsically asymmetric. On the source side, important evidence should not be omitted, since missing a crucial cue may directly harm downstream reasoning. On the target side, however, long videos contain many redundant or weakly relevant frames, which need not absorb mass. This asymmetry makes \emph{semi-unbalanced entropic optimal transport} a natural fit~\cite{chizat2018unbalanced,chizat2018scaling}: it preserves source-side evidence requirements while softly relaxing the target marginal, allowing mass to concentrate on evidence-bearing frames without artificially spreading it over redundant background content.

Based on this formulation, we instantiate a practical two-stage training-free pipeline, illustrated at the bottom of Figure~\ref{fig:intro}. In the first stage, we specialize diagonal-scaling updates for entropic semi-unbalanced OT~\cite{cuturi2013sinkhorn,chizat2018scaling,peyre2019computational} to derive \emph{OTFS-Sinkhorn}, which computes a globally consistent soft frame-importance distribution. In the second stage, we apply an exact dynamic program (DP) to extract a size-$k$ subset that optimizes the induced coverage-aware selection objective. Our contribution is not a new general theory of semi-unbalanced optimal transport itself. Rather, we show that long-video frame selection admits a principled transport-based formulation, and that this view leads to a clear and effective two-stage optimization pipeline.

Our formulation also clarifies the relationship to prior methods. In particular, when the source side collapses to a single evidence source, OTFS induces a Gibbs-type distribution over frames, making Q-Frame~\cite{zhang2025qframe}'s temperature-scaled distribution a one-dimensional special case of our framework. In contrast, AKS~\cite{tang2025adaptive} and DIG~\cite{li2025divide} are not exact instances of our objective; they are better understood as related staged approaches to budgeted frame selection. Experiments on three long-video understanding benchmarks show that OTFS consistently outperforms strong baselines.

Our main contributions are summarized as follows:
\begin{enumerate}
    \item We revisit long-video frame selection from a principled optimization perspective and formulate it as a structured evidence-allocation problem under limited visual budget.
    \item We instantiate this view with a semi-unbalanced optimal transport objective and develop a practical two-stage training-free pipeline consisting of OTFS-Sinkhorn for soft evidence allocation and an exact dynamic program for coverage-aware subset extraction.
    \item We show both analytically and empirically that this formulation is useful: in the single-source case, it induces a Gibbs-type distribution over frames, making Q-Frame's temperature-scaled distribution a direct one-dimensional special case of our framework; on long-video benchmarks, the full OTFS pipeline consistently outperforms strong baselines.
\end{enumerate}

\section{Related Work}

\paragraph{Video MLLMs and long-video modeling.}
Recent MLLMs have extended image-language models to video inputs and achieved strong progress in video understanding~\cite{lin2023videollava,cheng2024videollama,li2024llavaonevision,bai2025qwen3vl,zhang2025videollama3,zhu2025internvl3}. Long videos, however, remain difficult because their visual token length often exceeds the effective context budget of current models. Existing methods address this limitation through long-context training or transfer~\cite{zhang2024longcontexttransfer,liu2024kangaroo}, memory or streaming mechanisms~\cite{song2023moviechat,qian2024videostreaming}, compact and hierarchical video representations~\cite{jin2024chatunivi,weng2024longvlm,shen2024longvu,li2025videochatflash,xu2024slowfastllava,shi2025slowfastarch}, redundancy reduction~\cite{wang2024retake}, or retrieval-, tree-, and agent-style reasoning~\cite{wang2024videoagent,wang2024videotree}. These approaches mainly improve how video MLLMs process or organize long inputs. OTFS instead focuses on a complementary question: given a fixed downstream MLLM and a limited visual budget, how should frames be selected in a principled way?

\paragraph{Query-aware frame selection.}
More closely related to our setting, recent work studies selecting a small query-conditioned subset of frames before downstream reasoning. AKS~\cite{tang2025adaptive} combines query relevance with temporal coverage, Q-Frame~\cite{zhang2025qframe} samples from a temperature-scaled score distribution, FOCUS~\cite{zhu2026focus} formulates training-free keyframe search, and DIG~\cite{li2025divide} combines query-type-dependent routing with reward-based utility estimation. Other recent related methods either perform inference-time query-aware frame selection or temporal search~\cite{sun2025mdp3,liu2025bolt,ye2025rethinking,liang2024keyvideollm,yu2025framevoyager}, or enhance video question answering through query-conditioned visual encoding, modular grounding, caption aggregation, or grid-based video representations~\cite{ganz2024qavit,min2024morevqa,zhang2023simplellmvideoqa,kim2024imagegrid}.
 Most existing selectors still rely on scalar scoring and staged procedures, followed by top-$k$, sampling, routing, or hand-designed search rules. OTFS instead formulates frame selection as a principled evidence-allocation problem with an explicit optimization objective, and couples semi-unbalanced transport with exact coverage-aware subset extraction. This view also clarifies the relation to prior methods: Q-Frame~\cite{zhang2025qframe} is the single-source limit of our formulation, while AKS~\cite{tang2025adaptive} and DIG~\cite{li2025divide} are better understood as related staged selectors rather than exact solvers of a common objective.

\paragraph{Optimal transport for long-video understanding.}
Optimal transport (OT) has been explored for long-video modeling
and compression. Lin~\citep{lin2024multi}
uses OT to mitigate noisy clip--caption and frame--word
correspondences during video--language pre-training.
OTT-Vid~\citep{kang2026ott} applies OT between tokens in
neighboring frames to estimate temporal compressibility and
allocate token-level compression budgets.
In contrast, OTFS employs semi-unbalanced OT to integrate
heterogeneous evidence into query-aware frame importance.

\section{Method}
\label{sec:method}

\subsection{Overview}

Our goal is to select a subset of $k$ frames from a long video that preserves the visual evidence most useful for answering the input query. We refer to the proposed method as \textbf{OTFS}. The method consists of two optimization stages. First, we solve a semi-unbalanced optimal transport problem between heterogeneous evidence atoms induced by the query and auxiliary visual cues and the candidate video frames, yielding a soft global frame-importance distribution. Second, we convert this distribution into a size-$k$ subset through an exact dynamic program that accounts for temporal coverage.

This decomposition follows the structure of the problem: grounding heterogeneous evidence is a global allocation problem over all frames, whereas enforcing a fixed frame budget is a discrete subset optimization problem. OTFS handles these two roles separately through semi-unbalanced transport and exact combinatorial optimization.

\subsection{Semi-Unbalanced Evidence Allocation Formulation}
\label{subsec:formulation}

Let $q$ denote the input query and let $V=(x_1,\ldots,x_M)$ be an ordered sequence of video frames. A frame selector outputs an index set
\begin{equation}
S \subseteq \{1,\ldots,M\}, \qquad |S|=k,
\end{equation}
whose corresponding frames are passed to the downstream MLLM. The objective is to preserve, under the budget $k$, the visual evidence required to answer $q$.

Rather than reducing the query to a single frame-score vector, we introduce a set of \emph{evidence atoms}
\begin{equation}
\mathcal{Z} = \{z_1,\ldots,z_N\},
\end{equation}
where each atom denotes an evidence requirement induced by the query or auxiliary visual cues. This abstraction allows heterogeneous evidence to be represented within a single optimization problem.

For each evidence atom $z_i$ and frame $x_j$, let $r_{ij} \in [0,1]$ denote their compatibility, with larger values indicating stronger support. We convert compatibility into transport cost by
\begin{equation}
\label{eq:cost}
C_{ij} = 1 - r_{ij},
\end{equation}
which defines a cost matrix $C \in \mathbb{R}^{N \times M}$.

We treat the evidence atoms as the source domain and the video frames as the target domain. Let $\pi \in \mathbb{R}_+^{N \times M}$ be a transport plan, where $\pi_{ij}$ denotes how much mass of atom $z_i$ is assigned to frame $x_j$. Let $\mathbf{a} \in \Delta^N$ be a prior distribution over evidence atoms and $\mathbf{b} \in \Delta^M$ a prior distribution over frames, where $\Delta^d = \{u \in \mathbb{R}_+^d : \sum_\ell u_\ell = 1\}$ denotes the probability simplex.

We define the evidence-allocation objective as
\begin{equation}
\label{eq:main}
\min_{\pi \in \mathbb{R}_+^{N \times M}}
\; \langle C, \pi \rangle - \varepsilon H(\pi)
+ \gamma \, \mathrm{KL}\!\left(\pi^\top \mathbf{1} \,\|\, \mathbf{b}\right)
\quad
\text{s.t.}
\quad
\pi \mathbf{1} = \mathbf{a},
\end{equation}
where $\langle C,\pi\rangle$ denotes the Frobenius inner product and
\begin{equation}
H(\pi) = -\sum_{i=1}^N \sum_{j=1}^M \pi_{ij}(\log \pi_{ij} - 1)
\end{equation}
is the entropy regularizer. Here, $\varepsilon>0$ is the entropic regularization coefficient, and $\gamma$ controls the strength of the soft target-marginal regularization: larger $\gamma$ keeps the target marginal closer to $\mathbf{b}$, whereas smaller $\gamma$ allows mass to concentrate on evidence-supporting frames.

Equation~\eqref{eq:main} is a semi-unbalanced optimal transport formulation~\cite{cuturi2013sinkhorn,chizat2018unbalanced,chizat2018scaling,peyre2019computational}. The source marginal is enforced exactly, while the target marginal is only softly regularized. This asymmetry matches long-video frame selection: important evidence should not be discarded, whereas redundant or weakly relevant frames need not absorb mass. Given the optimal plan $\pi^{\star}$, we define the frame-importance distribution as the target marginal
\begin{equation}
\label{eq:weights}
\mathbf{w} = (\pi^{\star})^\top \mathbf{1},
\qquad
w_j = \sum_{i=1}^N \pi^{\star}_{ij}.
\end{equation}

\subsection{OTFS-Sinkhorn and the Single-Source Reduction}
\label{subsec:optimization}

Eq.~\eqref{eq:main} instantiates entropic semi-unbalanced OT for the proposed evidence-allocation problem, where evidence atoms form the source side and video frames form the KL-relaxed target side. Specializing the standard scaling characterization of semi-unbalanced OT~\cite{chizat2018unbalanced,chizat2018scaling,peyre2019computational} to this setting yields a simple solver, which we call OTFS-Sinkhorn.

Define the Gibbs kernel
\begin{equation}
\label{eq:kernel}
K_{ij} = \exp(-C_{ij}/\varepsilon).
\end{equation}
By standard results for entropically regularized semi-unbalanced OT, the optimizer admits the scaling form
\begin{equation}
\label{eq:factor}
\pi^{\star} = \mathrm{Diag}(\mathbf{u}) \, K \, \mathrm{Diag}(\mathbf{v}),
\end{equation}
for positive scaling vectors $\mathbf{u} \in \mathbb{R}_{++}^N$ and $\mathbf{v} \in \mathbb{R}_{++}^M$. Under this factorization, the scaling vectors satisfy
\begin{align}
\label{eq:u}
\mathbf{u} &= \mathbf{a} \oslash (K\mathbf{v}), \\
\label{eq:v}
\mathbf{v} &= \left(\mathbf{b} \oslash (K^\top \mathbf{u})\right)^{\gamma/(\gamma+\varepsilon)},
\end{align}
where $\oslash$ denotes element-wise division.

These updates define a simple fixed-point solver, which we refer to as \emph{OTFS-Sinkhorn}. Starting from $\mathbf{v}^{(0)}=\mathbf{1}$, we iterate
\begin{align}
\mathbf{u}^{(t)} &\leftarrow \mathbf{a} \oslash (K\mathbf{v}^{(t-1)}), \\
\tilde{\mathbf{v}} &\leftarrow \mathbf{b} \oslash (K^\top \mathbf{u}^{(t)}), \\
\mathbf{v}^{(t)} &\leftarrow \tilde{\mathbf{v}}^{\,\gamma/(\gamma+\varepsilon)},
\end{align}
until convergence, after which $\pi^{\star}$ is recovered through Eq.~\eqref{eq:factor}. Relative to balanced Sinkhorn, the distinctive term is the exponent $\gamma/(\gamma+\varepsilon)$ in Eq.~\eqref{eq:v}, which arises from the KL-relaxed target marginal. A derivation specialized to our objective is provided in the appendix.

\paragraph{Single-source Gibbs reduction.}
The following special case is important for interpreting OTFS and relating it to prior query-aware selectors. Consider $N=1$, with source marginal $a_1=1$, cost row $c_j = C_{1j}$, and a uniform frame prior $b_j = 1/M$ for all $j$. Then Eq.~\eqref{eq:main} reduces to
\begin{equation}
\min_{\pi \in \Delta^M}
\sum_{j=1}^M \pi_j c_j
+ \varepsilon \sum_{j=1}^M \pi_j \log \pi_j
+ \gamma \sum_{j=1}^M \pi_j \log \frac{\pi_j}{b_j},
\end{equation}
whose unique solution is
\begin{equation}
\pi_j
=
\frac{\exp\!\left(-c_j/(\varepsilon+\gamma)\right)}
{\sum_{\ell=1}^M \exp\!\left(-c_\ell/(\varepsilon+\gamma)\right)}.
\end{equation}
The derivation is given in the appendix.

This result shows that one-dimensional frame scoring arises as a degenerate single-source limit of OTFS. In particular, the strategy in Q-Frame~\cite{zhang2025qframe} is a direct one-dimensional special case: under a single evidence atom, a uniform frame prior, and a cost induced from the same scalar relevance signal, OTFS yields exactly the same Gibbs-form distribution as Q-Frame's temperature-scaled sampling rule. By contrast, AKS~\cite{tang2025adaptive} and DIG~\cite{li2025divide} are not exact instances of Eq.~\eqref{eq:main}, but can still be understood within the same broader view. AKS combines scalar query relevance with coverage-aware budget allocation, while DIG augments scalar or sparse utility-style signals with query-type-dependent routing and additional extraction logic. In this sense, Q-Frame corresponds to the single-source limit of OTFS, whereas AKS and DIG are better understood as related heuristic or staged procedures that preserve parts of the same query-aware, budget-aware intuition without being derived as exact solvers of a common optimization objective.

\subsection{Instantiating the Evidence Atoms}
\label{subsec:instantiation}

The formulation itself is agnostic to how evidence atoms are constructed. In our experiments, we instantiate OTFS with three complementary evidence sources motivated by prior query-aware frame-selection methods~\cite{tang2025adaptive,zhang2025qframe,li2025divide}: semantic alignment to the query, answer-oriented utility, and temporal variation in the visual stream.

\paragraph{Semantic relevance.}
The first evidence source measures direct semantic alignment between the query $q$ and each frame $x_j$. Following prior query-aware selectors~\cite{tang2025adaptive,zhu2026focus}, we use the image-text matching score from BLIP ITM~\cite{li2022blip} and define
\begin{equation}
r_j^{\mathrm{sem}} = \mathrm{BLIP}_{\mathrm{ITM}}(q, x_j) \in [0,1].
\end{equation}

\paragraph{Utility reward.}
We also include a reward-based utility signal inspired by DIG~\cite{li2025divide}, which estimates the downstream usefulness of each frame for question answering. To keep the computation practical on long videos, we evaluate the reward only on a temporally sampled subset of frames and linearly interpolate the resulting scores to the remaining candidates. In practice, we follow the public DIG implementation to obtain the sampled frames and their corresponding reward values.

\paragraph{Temporal transition.}
The third evidence source captures temporal variation that may indicate event boundaries or visually salient state changes. Following DIG~\cite{li2025divide}, we compute DINOv2 features~\cite{oquab2023dinov2} for adjacent frames and define
\begin{equation}
r_j^{\mathrm{trans}} = 1 - \cos\bigl(\psi(x_j), \psi(x_{j-1})\bigr)
\end{equation}
for $j \ge 2$, where $\psi(x_j)$ denotes the DINOv2 feature of frame $x_j$.

Using these three sources, we instantiate OTFS with $N=3$ evidence atoms by setting
\begin{equation}
r_{1j}=r_j^{\mathrm{sem}}, \qquad
r_{2j}=r_j^{\mathrm{util}}, \qquad
r_{3j}=r_j^{\mathrm{trans}}.
\end{equation}
Before forming the transport objective, we normalize all three score sequences to $[0,1]$ so that they are comparable across evidence sources. The resulting compatibility matrix $R=(r_{ij})$ is converted into the transport cost matrix $C$ through Eq.~\eqref{eq:cost}. More generally, OTFS can incorporate additional evidence atoms without changing the optimization framework itself.

\subsection{Exact Coverage-Aware Budgeted Subset Extraction}

As defined in Eq.~\eqref{eq:weights}, the optimal transport plan induces a frame-importance distribution $\mathbf{w}$. However, the final subset should not be determined by marginal importance alone. In long videos, high-importance frames often cluster around the same event, while downstream reasoning also benefits from sufficient temporal spread, as highlighted in~\cite{tang2025adaptive}. To avoid the quadratic complexity of jointly optimizing evidence matching and temporal coverage in the framework of optimal transport, we adopt a principled two-stage relaxation. We treat subset extraction as a second-stage discrete optimization problem, using an exact dynamic program to enforce the frame budget and temporal coverage over the continuous OT marginal.

For an ordered subset $S=\{t_1 < \cdots < t_k\}$, we define the subset-extraction objective as maximizing
\begin{equation}
\label{eq:subset}
\mathcal{J}(S)
=
\sum_{\ell=1}^k w_{t_\ell}
+
\lambda \sum_{\ell=2}^k G(t_{\ell-1}, t_\ell),
\qquad
G(i,j) = \min(j-i,\delta),
\end{equation}
where $\delta > 0$ denotes the desired minimum temporal gap between two consecutively selected frames. The coverage reward $G(i,j)$ increases with temporal distance and saturates once the gap reaches $\delta$, encouraging temporal spread without disproportionately favoring very distant frame pairs.

Because Eq.~\eqref{eq:subset} has a chain-structured pairwise form, the globally optimal size-$k$ subset under this objective can be recovered exactly by dynamic programming. Let $\mathrm{DP}[i,c]$ denote the maximum objective value achievable by selecting $c$ frames from $\{1,\ldots,i\}$ with frame $i$ fixed as the last selected frame. Then
\begin{equation}
\label{eq:dp}
\mathrm{DP}[i,c]
=
w_i + \max_{j<i} \left( \mathrm{DP}[j,c-1] + \lambda G(j,i) \right),
\end{equation}
with initialization $\mathrm{DP}[i,1]=w_i$. The optimal subset is obtained by backtracking from $\max_i \mathrm{DP}[i,k]$. 

\subsection{Algorithm Summary}
\label{subsec:algorithm}

The complete OTFS pipeline is summarized in Algorithm~\ref{alg:otfs}. Given a query and a long video, we first compute the evidence sources, construct the transport cost matrix, solve the semi-unbalanced OT problem via OTFS-Sinkhorn, and obtain  frame-importance distribution from the target marginal. We then solve the coverage-aware dynamic program to recover the final subset of $k$ frames.

\begin{algorithm}[t]
\caption{OTFS for Long-Video Frame Selection}
\label{alg:otfs}
\begin{algorithmic}[1]
\REQUIRE Query $q$, frames $V=(x_1,\ldots,x_M)$, frame budget $k$, marginals $\mathbf{a},\mathbf{b}$, transport hyperparameters $\varepsilon,\gamma$
\STATE Compute semantic scores $r_j^{\mathrm{sem}}$, utility scores $r_j^{\mathrm{util}}$, and transition scores $r_j^{\mathrm{trans}}$
\STATE Construct $R=(r_{ij})$ and $C=(1-r_{ij})$
\STATE Form $K=\exp(-C/\varepsilon)$ and initialize $\mathbf{v}^{(0)}=\mathbf{1}$
\FOR{$t=1,\ldots,T$}
    \STATE $\mathbf{u}^{(t)} \leftarrow \mathbf{a} \oslash (K\mathbf{v}^{(t-1)})$
    \STATE $\tilde{\mathbf{v}} \leftarrow \mathbf{b} \oslash (K^\top \mathbf{u}^{(t)})$
    \STATE $\mathbf{v}^{(t)} \leftarrow \tilde{\mathbf{v}}^{\,\gamma/(\gamma+\varepsilon)}$
\ENDFOR
\STATE Recover $\pi^{\star}=\mathrm{Diag}(\mathbf{u}^{(T)})K\mathrm{Diag}(\mathbf{v}^{(T)})$
\STATE Compute frame weights $\mathbf{w}=(\pi^{\star})^\top \mathbf{1}$
\STATE Solve Eq.~\eqref{eq:subset} by dynamic programming and backtracking to obtain $S^{\star}$
\RETURN $S^{\star}$
\end{algorithmic}
\end{algorithm}

\section{Experiments}

\subsection{Experimental Settings}

\paragraph{Datasets.}
We evaluate OTFS on three representative long-video understanding benchmarks: Video-MME~\cite{fu2025videomme}, LongVideoBench~\cite{wu2024longvideobench}, and MLVU~\cite{zhou2024mlvu}. Their video durations range from a few minutes to about one hour, covering diverse long-video settings. Following the official evaluation protocols, we report accuracy on the standard evaluation splits. For each downstream MLLM, all methods are compared under the same frame budget within the LMMs-Eval~\cite{zhang2025lmms} framework. We do not use subtitle information in any experiment. Additional dataset statistics are deferred to the appendix.

\paragraph{Implementation details.}
We instantiate the selector on top of two video-capable MLLM backbones, Qwen2.5-VL-7B~\cite{bai2025qwen25vl} and Qwen3-VL-4B~\cite{bai2025qwen3vl}, and evaluate each method under two budgets, $k \in \{32,64\}$. Unless otherwise specified, all compared methods use the same prompt format, decoding setup, and downstream evaluation protocol, so performance differences are attributable to the frame selector itself. For OTFS, we use three evidence atoms corresponding to semantic relevance, utility reward, and temporal transition, with source prior $\mathbf{a} \propto (0.5,0.4,0.1)$ and a uniform frame-side prior $\mathbf{b}$ with $b_j = 1/M$ for $j=1,\ldots,M$. We set the transport hyperparameters to $\gamma=0.1$ and $\varepsilon=0.05$ in all experiments, and run OTFS-Sinkhorn for 100 iterations by default. Following AKS~\cite{tang2025adaptive}, we sample the input video at 1 fps to construct the candidate-frame stream. More details are provided in the appendix.

\subsection{Main Results}

We compare OTFS against uniform sampling and four representative training-free frame-selection methods: AKS~\cite{tang2025adaptive}, Q-Frame~\cite{zhang2025qframe}, FOCUS~\cite{zhu2026focus}, and DIG~\cite{li2025divide}. All baseline results are reproduced using their publicly released codebases under the same evaluation framework. Tables~\ref{tab:qwen25-main} and~\ref{tab:qwen3-main} report the results on the two downstream MLLM backbones.

OTFS achieves the strongest overall performance across both backbones and both frame budgets. It obtains the best result in 11 out of 12 backbone-budget-dataset settings, and in the remaining case is within 0.03 points of the strongest baseline. On Qwen2.5-VL-7B, OTFS improves over the best baseline by $1.11$, $1.64$, and $0.35$ points on Video-MME, LongVideoBench, and MLVU at $k=32$, and by $0.07$ and $0.43$ points on LongVideoBench and MLVU at $k=64$. On Qwen3-VL-4B, the gains are larger: OTFS surpasses the strongest baseline by $1.18$, $1.94$, and $3.83$ points at $k=32$, and by $0.48$, $0.45$, and $2.70$ points at $k=64$.

\begin{figure}[t]
  \centering
  \includegraphics[width=0.9\textwidth]{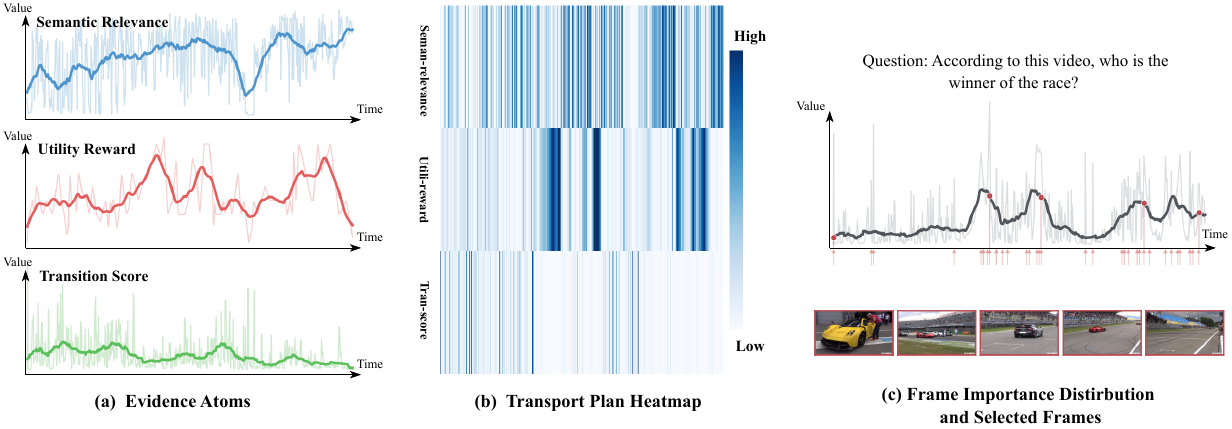}
  \caption{Left: Three evidence atoms over time. Middle: The optimal transport plan $\pi^*$, shown as a heatmap from evidence atoms to video frames. Different evidence sources assign mass to distinct temporal regions, highlighting their complementary roles in grounding the query. Right: The resulting frame-importance marginal $\mathbf{w}$ and the final subset selected by dynamic programming (red arrows indicate all selected frames, and red dots mark five representative examples).}
  \label{fig:visual}
\end{figure}

These results support the main claim of the paper. Compared with prior training-free baselines, OTFS benefits from formulating frame selection as a principled evidence-allocation problem with an explicit optimization objective, and from enforcing the final frame budget through exact subset extraction. The gains persist across datasets, backbones, and frame budgets, suggesting that the advantage of OTFS is structural rather than tied to a particular benchmark or model configuration. A qualitative visualization is provided in Figure~\ref{fig:visual}.

\begin{table}[t]
\centering
\small
\setlength{\tabcolsep}{5pt}
\caption{Main results with Qwen2.5-VL-7B.}
\label{tab:qwen25-main}
\resizebox{0.95\linewidth}{!}{%
\begin{tabular}{lcccccc}
\toprule
& \multicolumn{3}{c}{32 Frames} & \multicolumn{3}{c}{64 Frames} \\
\cmidrule(lr){2-4} \cmidrule(lr){5-7}
Method & Video-MME & LongVideoBench & MLVU & Video-MME & LongVideoBench & MLVU \\
\midrule
Uniform Sampling & 60.89 & 58.56 & 58.69 & 62.56 & 59.84 & 60.24 \\
Q-Frame~\cite{zhang2025qframe} & 60.77 & 59.01 & 60.89 & 62.92 & 60.88 & 63.64 \\
FOCUS~\cite{zhu2026focus} & 61.29 & 60.35 & 64.29 & 62.00 & 61.70 & 64.22 \\
DIG~\cite{li2025divide} & 62.11 & 61.85 & 65.49 & \textbf{64.25} & 63.28 & 66.43 \\
AKS~\cite{tang2025adaptive} & 62.44 & 62.83 & 65.39 & 63.81 & 63.87 & 64.96 \\
\midrule
OTFS (ours) & \textbf{63.55} & \textbf{64.47} & \textbf{65.84} & 64.22 & \textbf{63.94} & \textbf{66.86} \\
\bottomrule
\end{tabular}%
}
\end{table}

\begin{table}[t]
\centering
\small
\setlength{\tabcolsep}{5pt}
\caption{Main results with Qwen3-VL-4B.}
\label{tab:qwen3-main}
\resizebox{0.95\linewidth}{!}{%
\begin{tabular}{lcccccc}
\toprule
& \multicolumn{3}{c}{32 Frames} & \multicolumn{3}{c}{64 Frames} \\
\cmidrule(lr){2-4} \cmidrule(lr){5-7}
Method & Video-MME & LongVideoBench & MLVU & Video-MME & LongVideoBench & MLVU \\
\midrule
Uniform Sampling & 61.77 & 60.73 & 61.20 & 64.18 & 62.75 & 64.47 \\
Q-Frame~\cite{zhang2025qframe} & 62.48 & 61.55 & 63.33 & 65.22 & 64.32 & 66.84 \\
FOCUS~\cite{zhu2026focus} & 61.11 & 62.75 & 66.54 & 62.51 & 63.79 & 67.96 \\
DIG~\cite{li2025divide} & 63.22 & 64.02 & 66.97 & 65.48 & 66.19 & 67.56 \\
AKS~\cite{tang2025adaptive} & 63.96 & 65.37 & 68.27 & 65.55 & 66.71 & 69.27 \\
\midrule
OTFS (ours) & \textbf{65.14} & \textbf{67.31} & \textbf{72.10} & \textbf{66.03} & \textbf{67.16} & \textbf{71.97} \\
\bottomrule
\end{tabular}%
}
\end{table}

\subsection{Ablation Study}

The ablations in Table~\ref{tab:ablation} are conducted on Qwen3-VL-4B with $k=32$. We vary one component at a time while keeping the others fixed in order to isolate the contributions of three design choices in OTFS: multi-source evidence modeling, semi-unbalanced transport geometry, and exact budgeted subset extraction. Additional hyperparameter sensitivity results are deferred to the appendix.

\begin{table}[t]
\centering
\caption{Ablation study of OTFS on Qwen3-VL-4B with $k=32$.}
\label{tab:ablation}
\resizebox{0.8\linewidth}{!}{%
\begin{tabular}{lccc}
\toprule
Variant & Video-MME & LongVideoBench & MLVU \\
\midrule
\multicolumn{4}{l}{\textit{Evidence modeling}} \\
Semantic only                  & 63.88 & 65.14 & 69.06 \\
Semantic + Utility            & 65.03 & \textbf{67.46} & 71.79 \\
Semantic + Utility + Transition (full) & \textbf{65.14} & 67.31 & \textbf{72.10} \\
\midrule
\multicolumn{4}{l}{\textit{Transport geometry}} \\
Tuned score fusion        & 64.18 & 65.81 & 70.23 \\
Balanced OT                   & 64.62 & 65.59 & 70.42 \\
Semi-unbalanced OT (full)     & \textbf{65.14} & \textbf{67.31} & \textbf{72.10} \\
\midrule
\multicolumn{4}{l}{\textit{Subset extraction}} \\
Top-$k$                       & 63.74 & 66.49 & 71.60 \\
Gumbel-Max sampling           & 64.62
& 65.89 & 71.09 \\
Dynamic programming (full)    & \textbf{65.14} & \textbf{67.31} & \textbf{72.10} \\
\bottomrule
\end{tabular}%
}
\end{table}

\paragraph{Effect of multi-source evidence modeling.}
We first examine how the choice of evidence sources affects performance. Using semantic relevance alone already gives a competitive selector, confirming that direct query-frame alignment is a strong signal. Adding the utility reward further improves performance on all three benchmarks, indicating that semantic alignment and downstream answer usefulness are complementary rather than interchangeable. Incorporating temporal transition yields the strongest overall variant, especially on Video-MME and MLVU, which suggests that temporal variation provides additional evidence not captured by query-conditioned scoring alone. Overall, the full three-source instantiation performs best because it combines semantic, utility, and temporal cues within the same global evidence-allocation objective.

\paragraph{Effect of transport geometry.}
We isolate the effect of the transport formulation by keeping the evidence atoms and DP extractor fixed. The tuned-fusion baseline uses the same three normalized evidence atoms and the same DP extractor as OTFS, but replaces transport with a grid-searched scalar weighted sum. We also compare with balanced OT, which enforces hard marginal constraints on both evidence and frame sides. Semi-unbalanced OT achieves the strongest overall performance, indicating that the gains are not merely due to multi-signal fusion or the DP extractor. Unlike scalar fusion, which collapses heterogeneous cues into a single frame score before selection, OTFS preserves source-level evidence structure through the transport plan and then induces a target-side frame-importance marginal. This better matches long-video frame selection, where important evidence should be preserved while redundant frames need not absorb mass.

\paragraph{Effect of the final subset extractor.}
Finally, we study the role of the discrete subset extractor. With the transport-induced frame weights fixed, we compare the proposed dynamic programming solver against deterministic top-$k$ selection and Gumbel-Max sampling~\cite{jang2016categorical}. Dynamic programming performs best on all three benchmarks. This result shows that final performance depends not only on the transport-derived frame-importance distribution, but also on how the frame budget is enforced. Exact optimization of the coverage-aware extraction objective yields better subsets than heuristic ranking or stochastic sampling.

\subsection{Complexity Analysis}
\label{complexity}

Let $N$ denote the number of evidence atoms, $M$ the number of candidate video frames, $T$ the number of OTFS-Sinkhorn iterations, and $k$ the final frame budget. The transport stage performs iterative matrix scaling on an $N \times M$ transport matrix and therefore costs $\mathcal{O}(TNM)$. The subsequent dynamic program has complexity $\mathcal{O}(M^2k)$ under the recurrence in Eq.~\eqref{eq:dp}. Hence the overall complexity of OTFS is
\begin{equation}
\mathcal{O}(TNM + M^2 k).
\end{equation}

In our default instantiation, $N=3$, so the transport stage scales linearly with the number of candidate frames. The DP stage has quadratic dependence on $M$ in its naive form, but it operates only on frame-level weights and a simple chain-structured coverage term, making its overhead modest relative to downstream MLLM inference. Moreover, the DP implementation can be accelerated to $\mathcal{O}(Mk)$ for this saturated-gap reward by maintaining prefix maxima.

Overall, once the evidence scores are available, the transport-and-extraction module adds only a small additional cost beyond evidence extraction. Additional efficiency analysis is deferred to the appendix.

\subsection{Robustness to Corrupted Evidence}

The performance of OTFS is ultimately bounded by the quality of the underlying evidence extractors. However, this does not necessarily mean that a local error in one evidence source will be directly or disproportionately amplified into the final frame subset.

To directly quantify this behavior, we independently corrupt each evidence source by mixing its atom-score vector with uniform random noise:
\begin{equation}
\widetilde{\mathbf r}^{(s)}
=(1-\rho)\mathbf r^{(s)}
+
\rho\boldsymbol{\xi},
\qquad
\boldsymbol{\xi}\sim\mathcal{U}(0,1),
\end{equation}
where $\rho \in \{0, 0.1, 0.3, 0.5\}$ controls the corruption strength, and $\rho=0$ corresponds to the original noise-free setting. We apply the corruption separately to the semantic, utility, and transition sources while keeping the other two sources unchanged.

\begin{table}[t]
\centering
\small
\caption{Robustness of OTFS to source noise on Qwen3-VL-4B with $k=32$.}
\label{tab:robustness}
\resizebox{0.6\linewidth}{!}{%
\begin{tabular}{lcccc}
\toprule
Source & Noise ratio $\rho$ & Video-MME & LongVideoBench & MLVU \\
\midrule
\multirow{4}{*}{Semantic}
& 0.0 & 65.14 & 67.31 & 72.10 \\
& 0.1 & 65.14 & 67.09 & 71.47 \\
& 0.3 & 64.37 & 66.19 & 70.62 \\
& 0.5 & 64.04 & 67.09 & 70.16 \\
\midrule
\multirow{4}{*}{Utility}
& 0.0 & 65.14 & 67.31 & 72.10 \\
& 0.1 & 65.63 & 67.46 & 72.15 \\
& 0.3 & 64.37 & 67.84 & 72.50 \\
& 0.5 & 64.52 & 66.12 & 71.59 \\
\midrule
\multirow{4}{*}{Transition}
& 0.0 & 65.14 & 67.31 & 72.10 \\
& 0.1 & 65.11 & 67.31 & 71.42 \\
& 0.3 & 64.96 & 67.17 & 71.41 \\
& 0.5 & 65.07 & 67.09 & 71.06 \\
\bottomrule
\end{tabular}%
}
\end{table}

The overall results in Table~\ref{tab:robustness} show limited sensitivity under low-to-moderate corruption and clearer degradation under severe corruption. OTFS is not equivalent to directly ranking frames by a single noisy score. In the transport stage, each evidence source contributes a fixed total mass determined by its source prior, preventing an individual source from contributing unrestricted weight to the final frame marginal. Entropic and marginal regularization further smooth the allocation over candidate frames. The subsequent DP stage then selects frames under the cardinality and temporal-coverage constraints, rather than directly taking the frames with the largest atom scores. Therefore, errors in individual atoms can affect the final subset, but they are not propagated through an unconstrained one-to-one mapping.

\section{Conclusion}

We introduced \textbf{OTFS}, a training-free and principled framework that formulates long-video frame selection as a semi-unbalanced optimal transport problem from heterogeneous evidence atoms to candidate video frames. By grounding frame importance in an explicit evidence-allocation objective, OTFS provides a structured alternative to prior hand-designed sampling procedures under the redundancy structure of long videos. Our formulation also clarifies how temperature-scaled one-dimensional scoring emerges as a single-source special case, placing several prior methods within a broader perspective. To obtain the final subset, OTFS couples transport-based evidence allocation with an exact dynamic program for coverage-aware budgeted extraction. Empirically, this combination yields consistent improvements across multiple long-video benchmarks, downstream MLLM backbones, and frame-budget regimes. We hope OTFS can serve as a useful step toward more principled visual compression and evidence selection for long-context multimodal reasoning.

\begin{ack}
\textbf{Funding.}
This work was supported by internal computational
resources provided by Yizhun Medical AI Co., Ltd.
No external funding was received for this research.

\textbf{Competing interests.}
We declare no additional competing
financial interests beyond the disclosed
institutional affiliations.
\end{ack}

\clearpage

\bibliographystyle{plainnat}
\bibliography{ref}


\clearpage

\appendix

\begin{center}
  {\LARGE \textbf{Appendix}}
\end{center}
  
The appendix is organized as follows:
\begin{itemize}
  \item Section~A provides additional theoretical results and proofs for the semi-unbalanced OT formulation and the
dynamic-programming-based subset selector.
  \item Section~B presents additional dataset details used in our experiments.
  \item Section~C reports hyperparameter sensitivity analyses.
  \item Section~D provides additional efficiency analysis.
  \item Section~E discusses the two-stage OT–DP decomposition.
  \item Section~F discusses the limitations of the current framework.
  \item Section~G discusses broader impacts.
\end{itemize}

\section{Additional Theoretical Results and Proofs}
\label{sec:proof}

In this appendix, we provide the derivations and proofs underlying the OTFS formulation. We first specialize the standard scaling characterization of entropic semi-unbalanced optimal transport~\cite{cuturi2013sinkhorn,chizat2018unbalanced,chizat2018scaling,peyre2019computational} to our evidence-allocation objective, which yields the OTFS-Sinkhorn updates used in the main text. We then analyze the task-specific consequence of this formulation: the single-source Gibbs reduction, which clarifies the connection between OTFS and one-dimensional frame-scoring rules. We also discuss the optimality of the dynamic-programming subset selector for the proposed coverage-aware extraction objective. Firstly, recall the optimization problem:
\begin{equation}
\label{eq:app-main}
\min_{\pi \in \mathbb{R}_+^{N \times M}}
\;
\langle C, \pi \rangle
- \varepsilon H(\pi)
+ \gamma \, \mathrm{KL}\!\left(\pi^\top \mathbf{1} \,\|\, \mathbf{b}\right)
\quad
\text{s.t.}
\quad
\pi \mathbf{1} = \mathbf{a},
\end{equation}
where
\[
H(\pi)
=
-
\sum_{i=1}^N \sum_{j=1}^M
\pi_{ij}(\log \pi_{ij} - 1).
\]

\subsection{Existence and Uniqueness}

Assume $\varepsilon > 0$, $\gamma > 0$, $C_{ij} < \infty$ for all $i,j$, and $\mathbf{a} \in \Delta^N$, $\mathbf{b} \in \Delta^M$ have strictly positive entries. Under these conditions, the optimization problem in Eq.~\eqref{eq:app-main} admits a unique minimizer.

To see this, define the feasible set
\[
\mathcal{U}
=
\left\{
\pi \in \mathbb{R}_+^{N \times M}
:
\pi \mathbf{1} = \mathbf{a}
\right\}.
\]
Since $\mathbf{a} \in \Delta^N$, the set $\mathcal{U}$ is nonempty, closed, and convex. Moreover, the row-sum constraint and nonnegativity imply that $\mathcal{U}$ is bounded, hence compact.

The objective in Eq.~\eqref{eq:app-main} consists of three parts. The transport cost $\langle C,\pi\rangle$ is linear in $\pi$. The negative entropy term $-\varepsilon H(\pi)$ is strictly convex on $\mathbb{R}_+^{N \times M}$ when $\varepsilon > 0$. The KL term
\[
\mathrm{KL}(\pi^\top \mathbf{1}\,\|\,\mathbf{b})
\]
is convex in the frame marginal $\pi^\top \mathbf{1}$ and therefore convex in $\pi$. Since $\mathbf{b}$ has strictly positive entries, this term is well defined over the feasible set.

Therefore, the overall objective is strictly convex on the convex feasible set $\mathcal{U}$, which implies uniqueness of the minimizer. Existence follows from lower semicontinuity of the objective together with compactness of $\mathcal{U}$.

\subsection{Scaling Form of the Optimal Plan}

We next specialize the standard diagonal-scaling form for entropically regularized semi-unbalanced OT~\cite{cuturi2013sinkhorn,chizat2018unbalanced,chizat2018scaling,peyre2019computational} to our objective.

Let
\[
K_{ij} = \exp(-C_{ij}/\varepsilon).
\]
Then the unique minimizer $\pi^\star$ of Eq.~\eqref{eq:app-main} admits the factorized form
\[
\pi^\star
=
\mathrm{Diag}(\mathbf{u})\, K\, \mathrm{Diag}(\mathbf{v})
\]
for some positive vectors $\mathbf{u} \in \mathbb{R}_{++}^N$ and $\mathbf{v} \in \mathbb{R}_{++}^M$.

This factorization follows directly from the first-order optimality conditions. Introduce Lagrange multipliers $\alpha_1,\dots,\alpha_N$ for the row constraints $\pi \mathbf{1} = \mathbf{a}$. The Lagrangian is
\[
\mathcal{L}(\pi,\alpha)
=
\sum_{i,j} \pi_{ij} C_{ij}
+
\varepsilon \sum_{i,j} \pi_{ij}\log \pi_{ij}
-
\varepsilon \sum_{i,j} \pi_{ij}
+
\gamma \, \mathrm{KL}(\pi^\top \mathbf{1}\,\|\,\mathbf{b})
+
\sum_i \alpha_i
\left(
a_i - \sum_j \pi_{ij}
\right).
\]
Let
\[
w_j = \sum_i \pi_{ij}.
\]
Differentiating $\mathcal{L}$ with respect to $\pi_{ij}$ gives
\[
C_{ij}
+
\varepsilon \log \pi_{ij}
+
\gamma \log \frac{w_j}{b_j}
-
\alpha_i
=
0.
\]
Rearranging,
\[
\pi_{ij}
=
\exp(\alpha_i/\varepsilon)
\exp(-C_{ij}/\varepsilon)
\exp\!\left(
-\frac{\gamma}{\varepsilon}
\log \frac{w_j}{b_j}
\right).
\]
Define
\[
u_i = \exp(\alpha_i/\varepsilon),
\qquad
v_j =
\exp\!\left(
-\frac{\gamma}{\varepsilon}
\log \frac{w_j}{b_j}
\right).
\]
Then
\[
\pi_{ij} = u_i K_{ij} v_j,
\]
which yields the stated diagonal-scaling form.

\subsection{Fixed-Point Equations and OTFS-Sinkhorn}

Under the factorization above, the corresponding scaling vectors satisfy the fixed-point equations
\[
\pi^\star
=
\mathrm{Diag}(\mathbf{u})\, K\, \mathrm{Diag}(\mathbf{v}),
\qquad
K=\exp(-C/\varepsilon).
\]
\begin{align}
\mathbf{u}
&=
\frac{\mathbf{a}}{K\mathbf{v}},
\label{eq:app-u}
\\
\mathbf{v}
&=
\left(
\frac{\mathbf{b}}{K^\top \mathbf{u}}
\right)^{\frac{\gamma}{\gamma+\varepsilon}},
\label{eq:app-v}
\end{align}
where all operations are elementwise. These are the standard scaling updates for this class of semi-unbalanced entropic OT problems and directly motivate the OTFS-Sinkhorn solver used in the main text.

The first equation comes from the hard source-side constraint:
\[
\pi^\star \mathbf{1}
=
\mathrm{Diag}(\mathbf{u}) K \mathbf{v}
=
\mathbf{a},
\]
which immediately gives Eq.~\eqref{eq:app-u}. For the second equation, let
\[
\mathbf{w}
=
(\pi^\star)^\top \mathbf{1}
=
\mathrm{Diag}(\mathbf{v}) K^\top \mathbf{u},
\]
so that $w_j = v_j (K^\top \mathbf{u})_j$. From the stationarity condition derived above,
\[
v_j
=
\left(
\frac{b_j}{w_j}
\right)^{\gamma/\varepsilon}.
\]
Substituting $w_j = v_j (K^\top \mathbf{u})_j$ yields
\[
v_j
=
\left(
\frac{b_j}{v_j (K^\top \mathbf{u})_j}
\right)^{\gamma/\varepsilon}.
\]
Equivalently,
\[
v_j^{1+\gamma/\varepsilon}
=
\left(
\frac{b_j}{(K^\top \mathbf{u})_j}
\right)^{\gamma/\varepsilon}.
\]
Taking both sides to the power $\varepsilon/(\gamma+\varepsilon)$ gives
\[
v_j
=
\left(
\frac{b_j}{(K^\top \mathbf{u})_j}
\right)^{\frac{\gamma}{\gamma+\varepsilon}},
\]
which is Eq.~\eqref{eq:app-v}.

\subsection{Single-Source Gibbs Reduction}

We now record an elementary single-source specialization of Eq.~\eqref{eq:app-main}. This reduction is included for interpretive purposes: it makes explicit how temperature-scaled one-dimensional frame scoring arises from our objective in a degenerate regime.

Consider the single-source case $N=1$, with source marginal $a_1=1$, cost row
\[
c = (c_1,\dots,c_M),
\]
and frame prior $\mathbf{b} \in \Delta^M$. Then the transport plan is simply a row vector
\[
\pi = (\pi_1,\dots,\pi_M),
\qquad
\sum_{j=1}^M \pi_j = 1,
\]
so Eq.~\eqref{eq:app-main} reduces to
\begin{equation}
\label{eq:single_app}
\min_{\pi \in \Delta^M}
\;
\sum_{j=1}^M \pi_j c_j
+
\varepsilon \sum_{j=1}^M \pi_j \log \pi_j
+
\gamma \sum_{j=1}^M \pi_j \log \frac{\pi_j}{b_j}.
\end{equation}

Expanding the KL term gives
\[
\gamma \sum_{j=1}^M \pi_j \log \frac{\pi_j}{b_j}
=
\gamma \sum_{j=1}^M \pi_j \log \pi_j
-
\gamma \sum_{j=1}^M \pi_j \log b_j.
\]
Hence the optimizer remains exponential in the negative cost, modulated by the prior $\mathbf{b}$; in this sense, the solution is Gibbs-type.

If the frame prior is uniform, namely
\[
b_j = \frac{1}{M}, \qquad j=1,\dots,M,
\]
then
\[
\sum_{j=1}^M \pi_j \log \frac{\pi_j}{b_j}
=
\sum_{j=1}^M \pi_j \log \pi_j + \log M,
\]
because $\sum_{j=1}^M \pi_j = 1$. The additive constant $\gamma \log M$ does not affect the optimizer, so Eq.~\eqref{eq:single_app} becomes equivalent to
\[
\min_{\pi \in \Delta^M}
\;
\sum_{j=1}^M \pi_j c_j
+
(\varepsilon+\gamma)\sum_{j=1}^M \pi_j \log \pi_j.
\]
Introducing a Lagrange multiplier for the simplex constraint gives
\[
c_j + (\varepsilon+\gamma)(\log \pi_j + 1) + \tau = 0,
\]
which implies
\begin{equation}
\label{eq:gibbs_uniform_app}
\pi_j
=
\frac{\exp\!\left(-c_j/(\varepsilon+\gamma)\right)}
{\sum_{\ell=1}^M \exp\!\left(-c_\ell/(\varepsilon+\gamma)\right)}.
\end{equation}
If the KL term is removed, the same calculation yields the standard softmax
\[
\pi_j
=
\frac{\exp(-c_j/\varepsilon)}
{\sum_{\ell=1}^M \exp(-c_\ell/\varepsilon)}.
\]

This form immediately explains the connection to temperature-scaled frame scoring. If the single-source cost is defined by
\[
c_j = 1-r_j,
\]
where $r_j$ is a scalar frame score, then Eq.~\eqref{eq:gibbs_uniform_app} becomes
\[
\pi_j \propto \exp\!\left(r_j/(\varepsilon+\gamma)\right),
\]
because the shared factor $\exp(-1/(\varepsilon+\gamma))$ cancels during normalization. In the KL-free case, this further reduces to
\[
\pi_j \propto \exp(r_j/\varepsilon).
\]
Therefore, temperature-scaled frame scoring can be interpreted as a single-source specialization of Eq.~\eqref{eq:app-main}.

\subsection{Optimality of the Dynamic Programming Selector}

We now justify the recursion used in the second-stage coverage-aware subset selection.

Recall the objective
\begin{equation}
\label{eq:subset_app}
\mathcal{J}(S)
=
\sum_{\ell=1}^k w_{t_\ell}
+
\lambda \sum_{\ell=2}^k G(t_{\ell-1}, t_\ell),
\end{equation}
for an ordered subset
\[
S = \{t_1 < t_2 < \cdots < t_k\}.
\]

Define
\[
\mathrm{DP}[i,c]
\]
as the optimal value of Eq.~\eqref{eq:subset_app} when selecting $c$ frames from $\{1,\dots,i\}$ with frame $i$ selected as the last frame. Then the recursion
\[
\mathrm{DP}[i,c]
=
w_i
+
\max_{j<i}
\Bigl(
\mathrm{DP}[j,c-1] + \lambda G(j,i)
\Bigr)
\]
holds, with initialization
\[
\mathrm{DP}[i,1] = w_i.
\]
The recursion follows from optimal substructure. Any optimal subset contributing to $\mathrm{DP}[i,c]$ must end with some predecessor $j < i$ immediately before frame $i$. Its objective therefore decomposes into the best value for selecting $c-1$ frames ending at $j$, namely $\mathrm{DP}[j,c-1]$, plus the incremental contribution of appending frame $i$, namely $w_i + \lambda G(j,i)$. Therefore the best subset ending at $i$ is obtained by choosing the predecessor $j < i$ that maximizes
\[
\mathrm{DP}[j,c-1] + \lambda G(j,i),
\]
which yields the stated recurrence. For $c=1$, the only feasible subset ending at $i$ is $\{i\}$, whose value is $w_i$.

\section{Additional Dataset Details}

\paragraph{Video-MME.}
Video-MME~\cite{fu2025videomme} is a comprehensive benchmark for evaluating MLLMs on video analysis.
It contains 900 manually curated videos with 2,700 human-annotated question-answer pairs, covering
diverse domains and temporal regimes from short clips to long videos. In particular, its video
durations range from 11 seconds to 1 hour, spanning short-, medium-, and long-term video understanding. In this work, we only use the visual stream and do not provide subtitle information
to the model, so that the comparison focuses on frame selection under a visual-only setting.

\paragraph{LongVideoBench.}
LongVideoBench~\cite{wu2024longvideobench} is a long-context interleaved video-language benchmark
designed to evaluate detailed reasoning over long inputs. It contains 3,763 web-collected videos and
6,678 human-annotated multiple-choice questions in 17 fine-grained categories. The videos can be as long as one hour, making the benchmark particularly suitable for testing whether
a frame selector can preserve sparse but answer-critical evidence. We do not use subtitle information in our experiments.

\paragraph{MLVU.}
MLVU~\cite{zhou2024mlvu} is a multi-task long-video understanding benchmark covering diverse video
genres such as movies, surveillance footage, egocentric videos, cartoons, and game videos. It is
designed to evaluate several aspects of long-video understanding through nine task categories,
including both multiple-choice and open-ended settings. The dataset is split into a development set containing 2,593 questions and a test set with 509 questions. For evaluation purposes, we exclusively employ multiple-choice questions from the development set, while excluding all open-ended questions.

\section{Hyperparameter Sensitivity}

We analyze the sensitivity of OTFS to its transport and subset-selection hyperparameters on Qwen3-VL-4B with $k=32$. Results are reported in Tables~\ref{tab:hyper} and~\ref{tab:dp-hyper}.

\paragraph{Transport hyperparameters.}
Table~\ref{tab:hyper} studies the source prior $\mathbf{a}$, entropic regularization $\varepsilon$, and marginal relaxation strength $\gamma$. Overall, OTFS is fairly stable across a broad range of settings on both Video-MME and LongVideoBench. The default source prior $(0.5,0.4,0.1)$ gives the best or near-best performance, suggesting that a moderate emphasis on semantic and utility signals with a smaller transition prior provides a good balance. For $\varepsilon$, intermediate values perform best, while overly small or overly large regularization leads to slight degradation, consistent with the trade-off between overly sharp and overly diffuse transport plans. A similar trend is observed for $\gamma$: moderate relaxation works best, whereas overly large $\gamma$ weakens the desired evidence-preserving behavior of the semi-unbalanced formulation.

\paragraph{Dynamic-programming hyperparameters.} 
We set the trade-off coefficient $\lambda$ to $0.5$ on Video-MME and
to $0.005$ on LongVideoBench and MLVU. The reason is that Video-MME includes more global questions and therefore
benefits more from stronger temporal coverage, whereas LongVideoBench and MLVU contain more locally grounded questions,
for which an overly large $\lambda$ may over-disperse the selected frames and weaken local evidence preservation. The temporal gap parameter $\delta$ is fixed to $4$ for all datasets.
  
Table~\ref{tab:dp-hyper} studies the trade-off coefficient $\lambda$ and temporal gap $\delta$ in the DP stage. Setting $\lambda=0$, which removes the coverage term, consistently underperforms the default setting, confirming the importance of temporal coverage in the final subset selection. At the same time, excessively large $\lambda$ also hurts performance, indicating that over-emphasizing coverage may sacrifice evidence quality. The method is also reasonably robust to $\delta$, with $\delta=4$ performing best overall in our setup and nearby values remaining competitive.

Overall, these results suggest that OTFS does not rely on fragile hyperparameter tuning, and the default settings used in the main experiments lie in stable regions of the parameter space.

\begin{table}[H]
\centering
\small
\setlength{\tabcolsep}{3pt}
\caption{Hyperparameter sensitivity on Qwen3-VL-4B with $k=32$. Default settings are *.}
\label{tab:hyper}
\resizebox{\linewidth}{!}{%
\begin{tabular}{lccccccccccccc}
\toprule
& \multicolumn{4}{c}{Source prior $\mathbf{a}$}
& \multicolumn{4}{c}{Entropic regularization $\varepsilon$}
& \multicolumn{4}{c}{Marginal relaxation $\gamma$} \\
\cmidrule(lr){2-5} \cmidrule(lr){6-9} \cmidrule(lr){10-13}
Dataset
& $(0.4,0.5,0.1)$ & $(0.5,0.4,0.1)^*$ & $(0.6,0.3,0.1)$ & $(0.3,0.6,0.1)$
& $0.02$ & $0.05^*$ & $0.1$ & $0.2$
& $0.02$ & $0.05$ & $0.1^*$ & $0.2$ \\
\midrule
Video-MME & 64.92 & 65.14 & 64.81 & 64.77 & 64.96 & 65.14 & 64.85 & 65.25 & 65.00 & 65.00 & 65.14 & 64.48  \\
LongVideoBench & 67.31 & 67.31 & 66.86 & 67.01 & 66.79 & 67.31 & 66.94 & 65.96 & 67.31 & 66.79 & 67.31 & 66.41  \\
\bottomrule
\end{tabular}%
}
\end{table}

\begin{table}[H]
\centering
\small
\setlength{\tabcolsep}{3pt}
\caption{Sensitivity to dynamic-programming hyperparameters on Qwen3-VL-4B with $k=32$.}
\label{tab:dp-hyper}
\resizebox{0.7\linewidth}{!}{%
\begin{tabular}{lcccccc|cccc}
\toprule
& \multicolumn{6}{c|}{Trade-off coefficient $\lambda$}
& \multicolumn{4}{c}{Temporal gap $\delta$} \\
\cmidrule(lr){2-7} \cmidrule(lr){8-11}
Dataset
& $0$ & \textbf{$0.005$} & $0.01$ & $0.02$ & $0.05$ & $0.5$
& $2$ & \textbf{$4^*$} & $6$ & $8$ \\
\midrule
Video-MME & 63.74 & 65.25 & 65.18 & 64.88 & 65.00 & 65.14  & 64.11 & 65.14 & 64.22 & 65.11 \\
LongVideoBench & 66.49 & 67.31 & 66.41 & 66.11 & 65.44 & 64.77 & 67.09 & 67.31 & 65.22 & 66.04 \\
\bottomrule
\end{tabular}%
}
\end{table}

\section{Efficiency Analysis}
\label{sec:efficiency}
We analyze efficiency by separating the optimization core of OTFS from the full evidence-extraction pipeline. As discussed in Sec~\ref{complexity}, the OTFS solver itself is lightweight: the Sinkhorn-based transport step and the subsequent DP-based subset extraction introduce only a small additional overhead once the evidence scores are available. This is also confirmed empirically in Table~\ref{tab:efficiency_all}, where the standalone Sinkhorn+DP stage requires only 5.8, 9.2, and 16.1 minutes on LongVideoBench, MLVU, and Video-MME, respectively. The wall-clock time of full OTFS is mainly determined by evidence extraction, while the transport-and-extraction module contributes only a small fraction of the total cost.

To study a more practical operating point, we further reduce the candidate-frame rate from 1 fps to 0.5 fps. As shown in Table~\ref{tab:efficiency_all}, this substantially lowers the preprocessing cost, bringing the overall runtime much closer to AKS, while Table~\ref{tab:fps} shows that the corresponding performance drop remains modest across all three benchmarks. This result suggests that the main computational bottleneck of the current system is not the OT-based optimization core, but the cost of obtaining rich evidence signals at dense temporal resolution.

Overall, these results support two conclusions. First, OTFS does not introduce a heavy optimization bottleneck beyond evidence extraction. Second, the practical efficiency of the full pipeline can be improved further through cheaper evidence estimators or lower candidate-frame rates, without changing the OTFS formulation itself. All experiments are conducted on 8 NVIDIA 4090 GPUs.

\begin{table}[H]
\centering
\small
\setlength{\tabcolsep}{4.5pt}
\caption{Efficiency comparison on three datasets. We report standard LMM inference time (minutes) under different frame budgets and additional key frame selection time (minutes) for different methods.}
\label{tab:efficiency_all}
\resizebox{0.95\linewidth}{!}{%
\begin{tabular}{lcccccccc}
\toprule
& \multicolumn{2}{c}{LMM Inference Time} & \multicolumn{6}{c}{Key Frame Selection Time} \\
\cmidrule(lr){2-3} \cmidrule(lr){4-9}
Dataset & 32 Frames & 64 Frames & Q-Frame & AKS & DIG & OTFS (1 fps) & OTFS (0.5 fps) & Sinkhorn+DP \\
\midrule
LongVideoBench & 27.3 & 53.6 & 658.3 & 660.5 & 300.9 & 971.2 & 634.1 & 5.8 \\
MLVU           & 43.5 & 81.5 & 902.3 & 906.0 & 567.1 & 1487.6 & 1033.4 & 9.2 \\
Video-MME      & 48.2 & 94.7 & 1856.1 & 1861.3 & 789.7  & 2676.1 & 1743.2 & 16.1 \\
\bottomrule
\end{tabular}%
}
\end{table}

\begin{table}[H]
\centering
\small
\setlength{\tabcolsep}{5pt}
\caption{Results with different fps.}
\label{tab:fps}
\resizebox{0.55\linewidth}{!}{%
\begin{tabular}{lccc}
\toprule
& \multicolumn{3}{c}{32 Frames}\\
\cmidrule(lr){2-4} 
Method & Video-MME & LongVideoBench & MLVU \\
\midrule
OTFS (1 fps) & 65.14 & 67.31 & 72.10  \\
OTFS (0.5 fps) & 65.07 & 67.01 & 71.55 \\
\bottomrule
\end{tabular}%
}
\end{table}

\section{Discussion on the two-stage OT–DP decomposition.}

\paragraph{Why the OT stage does not explicitly include $k$.}
The decomposition is deliberate and separates two roles. The OT stage estimates
a global, budget-independent evidence allocation over all candidate frames,
whereas the DP stage imposes the specific cardinality constraint and temporal
structure required by the final frame budget. As a result, the same transport
allocation can be reused for different frame budgets, such as $k=32$ and
$k=64$, without recomputing the evidence scores or the transport plan.

A direct joint formulation would require binary support variables
$y_j\in\{0,1\}$, for example,
$$
\sum_{j=1}^{M}y_j=k,
\qquad
\pi_{ij}\leq a_i y_j,
$$
together with a temporal coverage reward defined over the selected support.
Such a formulation would become a mixed discrete–continuous optimization
problem. It would no longer retain the convex transport subproblem or admit
the efficient Sinkhorn scaling updates used by OTFS. The proposed
two-stage pipeline is therefore a tractable decomposition that separates global
evidence grounding from exact budgeted subset extraction.

\paragraph{Why the DP stage uses the target marginal.}
Let
$$
w_j=\sum_{i=1}^{N}\pi^{\star}_{ij}
$$
be the target marginal of the optimal transport plan. For the additive evidence
retention term used in our current subset objective, we have
$$
\sum_{j\in S}w_j
=
\sum_{j\in S}\sum_{i=1}^{N}\pi^{\star}_{ij}
=
\sum_{i=1}^{N}\sum_{j\in S}\pi^{\star}_{ij}.
$$
Therefore, under the current objective, the marginal $\mathbf{w}$ is sufficient
for measuring the total transport mass retained by the selected subset. The DP
does not need the full transport matrix to optimize this particular additive
quantity.

\section{Limitations}
\label{sec:limit}
OTFS provides a principled optimization framework for frame selection, but its empirical performance still depends on the quality and task coverage of the underlying evidence atoms. In our current instantiation, these atoms are derived from semantic relevance, utility reward, and temporal transition. If these estimators are noisy, biased, or fail to capture the information truly required by a question, the transport stage may propagate such errors into the final subset. This issue can be more pronounced for questions requiring abstract temporal reasoning or subtle evidence not well captured by the current evidence design.

In addition, OTFS optimizes a frame-selection objective rather than the downstream reasoning objective of the MLLM itself. Thus, the selected subset should be interpreted as optimal with respect to the proposed transport-and-extraction objective, rather than guaranteed to be globally optimal for final question-answering accuracy. Relatedly, while the OTFS optimization core is lightweight, the runtime of the full system is currently dominated by evidence computation. Improving evidence quality and efficiency remains an important direction for future work.

\section{Broader Impacts}
\label{sec:impact}

This work studies long-video frame selection for multimodal large language models. A more efficient and principled
selector can reduce the computational cost of long-context video understanding, which may improve the accessibility of
video-language systems and lower the cost of deploying them in research and industry. In this sense, OTFS may
contribute to more efficient multimodal reasoning over long videos.

At the same time, improved video understanding may also enable harmful applications, including large-scale
surveillance, privacy intrusion, or more effective analysis of sensitive personal media. In addition, frame selection
errors may systematically omit important evidence, which can be particularly problematic in high-stakes settings such
as education, healthcare, or safety monitoring. Since OTFS is a training-free method built on upstream evidence
signals, biases or failures in those signals may propagate into the selected subset and subsequently affect downstream
reasoning.


\newpage
\input{main_checklist}

\end{document}

%% file: main_checklist.tex
\section*{NeurIPS Paper Checklist}


\begin{enumerate}

\item {\bf Claims}
    \item[] Question: Do the main claims made in the abstract and introduction accurately reflect the paper's contributions and scope?
    \item[] Answer: \answerYes{} 
    \item[] Justification: The last few sentences in \textbf{Abstract}; the last paragraph of \textbf{Introduction}.

\item {\bf Limitations}
    \item[] Question: Does the paper discuss the limitations of the work performed by the authors?
    \item[] Answer: \answerYes{} 
    \item[] Justification: Appendix~\ref{sec:limit}

\item {\bf Theory assumptions and proofs}
    \item[] Question: For each theoretical result, does the paper provide the full set of assumptions and a complete (and correct) proof?
    \item[] Answer: \answerYes{} 
    \item[] Justification: Appendix~\ref{sec:proof}

    \item {\bf Experimental result reproducibility}
    \item[] Question: Does the paper fully disclose all the information needed to reproduce the main experimental results of the paper to the extent that it affects the main claims and/or conclusions of the paper (regardless of whether the code and data are provided or not)?
    \item[] Answer: \answerYes{} 
    \item[] Justification: The paper discloses the datasets, backbones, frame budgets, evidence sources, hyperparameters, evaluation framework, baselines, ablations, and additional sensitivity/efficiency analyses. This provides sufficient information to understand and reproduce the main experimental protocol.

\item {\bf Open access to data and code}
    \item[] Question: Does the paper provide open access to the data and code, with sufficient instructions to faithfully reproduce the main experimental results, as described in supplemental material?
    \item[] Answer: \answerNo{} 
    \item[] Justification: The paper uses public datasets and provides substantial implementation details, but anonymized code and full reproduction scripts are not included at submission time. We will release code upon acceptance.

\item {\bf Experimental setting/details}
    \item[] Question: Does the paper specify all the training and test details (e.g., data splits, hyperparameters, how they were chosen, type of optimizer) necessary to understand the results?
    \item[] Answer: \answerYes{} 
    \item[] Justification: The paper specifies the evaluation datasets, backbones, budgets, evidence sources, transport hyperparameters, frame sampling rate, and comparison protocol in the main paper, with additional dataset details and hyperparameter sensitivity analyses provided in the appendix.

\item {\bf Experiment statistical significance}
    \item[] Question: Does the paper report error bars suitably and correctly defined or other appropriate information about the statistical significance of the experiments?
    \item[] Answer: \answerNo{} 
    \item[] Justification: The experimental setup in this study involved extensive testing across multiple datasets and with various types of backbone networks, consistently demonstrating improvements through our method. This extensive validation across diverse conditions ensures the reproducibility and reliability of the results.

\item {\bf Experiments compute resources}
    \item[] Question: For each experiment, does the paper provide sufficient information on the computer resources (type of compute workers, memory, time of execution) needed to reproduce the experiments?
    \item[] Answer: \answerYes{} 
    \item[] Justification: The paper includes complexity analysis in the main paper, efficiency measurements and compute resources in appendix~\ref{sec:efficiency}.
    
\item {\bf Code of ethics}
    \item[] Question: Does the research conducted in the paper conform, in every respect, with the NeurIPS Code of Ethics \url{https://neurips.cc/public/EthicsGuidelines}?
    \item[] Answer: \answerYes{} 
    \item[] Justification: Our work does not violate the code of ethics.

\item {\bf Broader impacts}
    \item[] Question: Does the paper discuss both potential positive societal impacts and negative societal impacts of the work performed?
    \item[] Answer: \answerYes{} 
    \item[] Justification: Appendix~\ref{sec:impact} discusses the societal impacts.
    
\item {\bf Safeguards}
    \item[] Question: Does the paper describe safeguards that have been put in place for responsible release of data or models that have a high risk for misuse (e.g., pre-trained language models, image generators, or scraped datasets)?
    \item[] Answer: \answerNA{} 
    \item[] Justification: The paper does not release a high-risk generative model, scraped dataset, or other asset that would require special safeguards for controlled release. The work studies an inference-time frame-selection method built on existing public benchmarks and pretrained models.

\item {\bf Licenses for existing assets}
    \item[] Question: Are the creators or original owners of assets (e.g., code, data, models), used in the paper, properly credited and are the license and terms of use explicitly mentioned and properly respected?
    \item[] Answer: \answerYes{} 
    \item[] Justification: The datasets and codes used in the paper have been cited in their original literature.

\item {\bf New assets}
    \item[] Question: Are new assets introduced in the paper well documented and is the documentation provided alongside the assets?
    \item[] Answer: \answerNA{} 
    \item[] Justification: The code is not publicly available at this time. We will make them available later.
    \item[] Guidelines:

\item {\bf Crowdsourcing and research with human subjects}
    \item[] Question: For crowdsourcing experiments and research with human subjects, does the paper include the full text of instructions given to participants and screenshots, if applicable, as well as details about compensation (if any)? 
    \item[] Answer: \answerNA{} 
    \item[] Justification: This paper does not involve crowdsourcing nor research with human subjects.

\item {\bf Institutional review board (IRB) approvals or equivalent for research with human subjects}
    \item[] Question: Does the paper describe potential risks incurred by study participants, whether such risks were disclosed to the subjects, and whether Institutional Review Board (IRB) approvals (or an equivalent approval/review based on the requirements of your country or institution) were obtained?
    \item[] Answer: \answerNA{} 
    \item[] Justification: The paper does not involve human-subject research or new data collection from participants, so IRB approval is not applicable

\item {\bf Declaration of LLM usage}
    \item[] Question: Does the paper describe the usage of LLMs if it is an important, original, or non-standard component of the core methods in this research? Note that if the LLM is used only for writing, editing, or formatting purposes and does \emph{not} impact the core methodology, scientific rigor, or originality of the research, declaration is not required.
    \item[] Answer: \answerYes{} 
    \item[] Justification: LLM/VLM components are part of the experimental pipeline and evaluation setting, including the use of video-capable MLLM backbones for downstream evaluation and a reward-based utility signal obtained following the public DIG implementation. These usages are described in Sections 3.4 and 4.1.

\end{enumerate}